\documentclass{article}

\usepackage{microtype}
\usepackage{graphicx}
\usepackage{subfigure}
\usepackage{booktabs} % for professional tables
\usepackage{float}
\usepackage{placeins}
\usepackage[ruled,noend,vlined]{algorithm2e}
\usepackage{amsmath}

\usepackage{hyperref}

\usepackage[accepted]{icml2020}

\icmltitlerunning{Routing Networks for Continual Learning}

\begin{document}

\twocolumn[
\icmltitle{Routing Networks with Co-training for Continual Learning}

\begin{icmlauthorlist}
\icmlauthor{Mark Collier}{goo}
\icmlauthor{Efi Kokiopoulou}{goo}
\icmlauthor{Andrea Gesmundo}{goo}
\icmlauthor{Jesse Berent}{goo}
\end{icmlauthorlist}

\icmlaffiliation{goo}{Google AI, Zurich}

\icmlcorrespondingauthor{Mark Collier}{markcollier@google.com}

\icmlkeywords{Continual learning, routing networks, expert networks}

\vskip 0.3in
]



\printAffiliationsAndNotice{}  % leave blank if no need to mention equal contribution

\begin{abstract}
% Many continual learning methods can be characterized as either altering the learning algorithm in a fixed capacity neural network or dynamically growing the capacity of the network to handle new tasks. We propose to use \textit{fixed capacity} sparse routing networks for continual learning. We retain the advantages of architectural solutions to the continual learning problem, in that different paths through the network can be learned for different tasks. However, we stay within the regime of fixed capacity networks which are more realistic for real-world use cases. We find it is necessary to develop a new training method for routing networks, which we call \textit{co-training} which avoids poorly initialized experts when new tasks are presented. In initial experiments, when combined with a small episodic memory replay buffer, sparse routing networks with co-training outperform densely connected networks on the MNIST-Permutations and MNIST-Rotations benchmarks.

The core challenge with continual learning is catastrophic forgetting, the phenomenon that when neural networks are trained on a sequence of tasks they rapidly forget previously learned tasks. It has been observed that catastrophic forgetting is most severe when tasks are dissimilar to each other. We propose the use of sparse routing networks for continual learning. For each input, these network architectures activate a different path through a network of experts. Routing networks have been shown to \textit{learn} to route similar tasks to overlapping sets of experts and dissimilar tasks to disjoint sets of experts. In the continual learning context this behaviour is desirable as it minimizes interference between dissimilar tasks while allowing positive transfer between related tasks. In practice, we find it is necessary to develop a new training method for routing networks, which we call \textit{co-training} which avoids poorly initialized experts when new tasks are presented. When combined with a small episodic memory replay buffer, sparse routing networks with co-training outperform densely connected networks on the MNIST-Permutations and MNIST-Rotations benchmarks.

\end{abstract}

\section{Introduction}
\label{sec:intro}

Continual learning is the challenge of training machine learning models on a non-stationary stream of data that cannot be  stored in its entirety. Often it is assumed the stream of data consists of discrete tasks \cite{kirkpatrick2017overcoming}. The major issue to be addressed in continual learning is \textit{catastrophic forgetting} \cite{CatastrophicForgetting.89}, which occurs when a neural network has severely degraded performance on previously learned tasks as a result of learning a new task. In addition to avoiding catastrophic forgetting ideally we would like methods to enable positive forward \textit{and} backward transfer i.e.\ learning previous tasks helps with learning new tasks and learning new tasks improves performance on past tasks. Many continual learning methods rely on regularizing the loss function and/or modifying the learning procedure e.g. EWC \cite{kirkpatrick2017overcoming}, GEM \cite{lopez2017gradient}, MER \cite{riemer2019learning}. These methods are typically implemented on top of standard neural network architectures. On the other hand, there are a range of continual learning methods which propose new network architectures for continual learning \cite{rusu2016progressive,hu2019overcoming,aljundi2017expert}. A major limitation of these methods is that they typically add capacity to the network when a new task is introduced. This linear scaling of network capacity with the number of tasks, limits these methods' applicability in many real-world settings.

\begin{figure}[t]
% \vskip 0.2in
\centering
\centerline{\includegraphics[width=0.8\columnwidth]{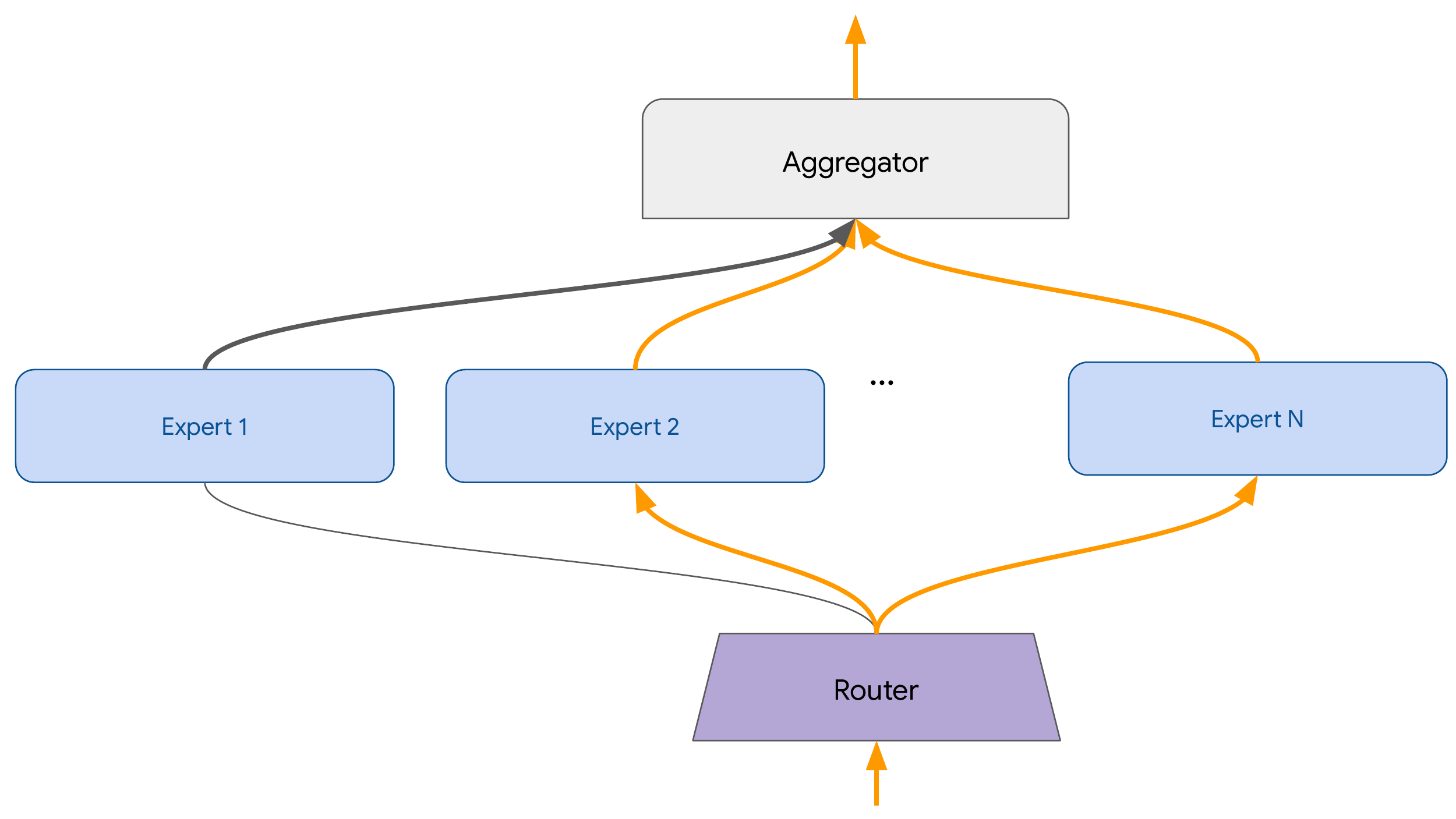}}
\caption{Router with N experts and generic aggregator.}
\label{fig:router}
% \vskip -0.2in
\end{figure}

Routing networks are a class of sparse neural network architectures. They consist of layers of experts, where each expert is an arbitrary neural network, see Fig.\ \ref{fig:router}. Associated with each layer is a router, which outputs a probability distribution over the experts to be activated for a given input. The router may itself be a neural network. In our work, the router is simply a matrix of routing probabilities conditioned on the task ID i.e.\ the $(i, j)^{th}$ element of the matrix is the probability of the $j^{th}$ expert being activated for task $i$. Each activated expert produces an output which is aggregated before being passed to the next layer. Typically this aggregation is simply a weighted sum of the expert outputs, with the routing probabilities as weights.

Sparsely-Gated Mixture-of-Experts \cite{shazeer2017outrageously} and other routing network variants \cite{rosenbaum2017routing,veit2018convolutional,maziarz2019gumbel,guo2019spottune} can be trained by gradient descent. These networks learn to route similar examples/tasks to the same experts, creating specialist experts resulting in improved predictive performance compared to non-sparse network architectures \cite{ramachandran2018diversity}. We propose the use of a fixed capacity architecture, sparse routing networks, for continual learning for the following reasons:

\begin{enumerate}
    \item Continual learning is inherently multi-task, routing networks have been shown to outperform densely connected networks for multi-task learning when the data is not presented continually \cite{rosenbaum2017routing,maziarz2019gumbel}.
    \item Sparse activations on the forward pass of the routing network implies sparse gradients. As a result if dissimilar tasks learn disjoint routes through the network then gradients from one task will not interfere with another task's weights, avoiding catastrophic forgetting. Likewise if similar tasks share a path through the network then this promotes positive transfer.
    \item Routing networks are fixed capacity networks, avoiding the linear scaling of capacity associated with other architectural solutions to continual learning.
    \item Routing networks are general purpose architectures, so can be combined with other continual learning methods e.g.\ EWC, MER, GEM, episodic memory.
\end{enumerate}

We find that direct application of routing networks to continual learning may not yield the desired results right away. We observe that when routing networks are trained continually,  some experts are not used by any tasks. When a new task is presented then the router must choose between a well trained expert and another expert close to its random initialization, often the router makes the greedy choice, leaving unused capacity and leading to interference between tasks. We propose the \textit{co-training} method to solve this problem; co-training presents examples from the current task to experts unused on any tasks, see \S \ref{sec:method} for a full description.

We combine the Sparsely-Gated Mixture-of-Experts, trained with co-training, with a simple but successful continual learning method, a small episodic memory replay buffer \cite{chaudhry2019continual}. On the MNIST-Permutations and MNIST-Rotations benchmarks we demonstrate improved average accuracy and reduced negative backward transfer. On MNIST-Rotations we have prior knowledge of which tasks are most related to each other. By examining the learned routing decisions we see that the routing networks learn similar notions of task similarity.

\section{Routing Networks with Co-training}
\label{sec:method}

\begin{algorithm}
\SetAlgoLined
 \textbf{Given:} network weights $\theta$, routing matrix $\mathcal{R}$, learning rate $\alpha$, co-training learning rate $\alpha_c$, layer $l$ max active experts $k_l$, used experts data structure $\mathcal{E}$ \\
 \textbf{Init:} empty episodic memory $\mathcal{M} \leftarrow \{\}$\\
 Set all experts in all layers to be unused $\mathcal{E}[l, t, e] \leftarrow \textrm{False}, \ \ \forall $ layers $l$, tasks $t$ and experts $e$ in layer $l$\\
 \For{$t$ = $1 ,..., T$}{
     \For{$(\mathbf{x}, y)$ in $\mathcal{D}_t$}{
      $\hat{\mathcal{R}} \leftarrow$ sample($\mathcal{R}$) \\
      $(\theta, \mathcal{R}) \leftarrow SGD(\mathbf{x}, y, \alpha, \theta \odot \hat{\mathcal{R}}, \mathcal{R})$\\

      // Draw sample from memory buffer \\
      $(\mathbf{x}_{\mathcal{M}}, y_{\mathcal{M}}) =$ sample\_batch($\mathcal{M}$) \\
      $\hat{\mathcal{R}} \leftarrow$ sample($\mathcal{R}$) \\
      $(\theta, \mathcal{R}) \leftarrow SGD(\mathbf{x}_{\mathcal{M}}, y_{\mathcal{M}}, \alpha, \theta \odot \hat{\mathcal{R}}, \mathcal{R})$\\
      
      $\mathcal{M} \leftarrow \mathcal{M} \cup (\mathbf{x}, y)$ // update episodic memory \\
      
      // Update unused experts \\
       \For{$l$ = $1 ,..., L$}{
         $\mathcal{E}[l, t, e] \leftarrow e \in {\textrm{arg} \ k_l \ } \mathcal{R}[l, t, :]$\\
       }
      
      // Co-training gradient step \\
      $(\theta, \_) \leftarrow SGD(\mathbf{x} \cup \mathbf{x}_{\mathcal{M}}, y \cup y_{\mathcal{M}}, \alpha_c, \theta \odot (1 - \mathcal{E}), \mathcal{R})$\\
      
  }
 }
\caption{Routing networks with co-training and reservoir sampling episodic memory for continual learning.}
\label{algo:method}
%  \vspace{-0.2in}
\end{algorithm}

\begin{table*}[tbh]
\caption{BWT and ACC for MNIST-Perm and MNIST-Rot. Averages $\pm$ one standard deviation over 15 runs are displayed for each metric.}
\label{tab:results}
\vskip 0.15in
\begin{center}
\begin{small}
\begin{sc}
\begin{tabular}{l|cc|cc|}
\toprule
\textbf{Method} & \multicolumn{2}{c|}{\textbf{MNIST-Perm}} & \multicolumn{2}{c|}{\textbf{MNIST-Rot}} \\
 & BWT & ACC & BWT & ACC \\
\midrule
Shared bottom & $-0.219 \pm 0.022$ & $0.726 \pm 0.021$ & $-0.269 \pm 0.025$ & $0.693 \pm 0.024$\\
Shared bottom + replay buffer & $-0.057 \pm 0.002$ & $0.912 \pm 0.002$ & $-0.057 \pm 0.004$ & $0.920 \pm 0.004$\\
\midrule
MoE & $-0.122 \pm 0.011$ & $0.839 \pm 0.011$ & $-0.250 \pm 0.044$ & $0.721 \pm 0.042$\\
MoE + replay buffer & $-0.040 \pm 0.004$ & $0.918 \pm 0.003$ & $-0.038 \pm 0.004$ & $0.923 \pm 0.004$\\
MoE + replay buffer + co-training & $\mathbf{-0.038 \pm 0.003}$ & $\mathbf{0.920 \pm 0.002}$ & $\mathbf{-0.034 \pm 0.005}$ & $\mathbf{0.929 \pm 0.004}$\\
\bottomrule
\end{tabular}
\end{sc}
\end{small}
\end{center}
\vskip -0.1in
\end{table*}

There is a challenge with directly applying routing networks to continual learning. When trained in a non-continual multi-task fashion, examples are drawn at random from all tasks at each training step. This aids diversity in the routing decisions, encouraging the use of all the experts for all the tasks collectively. However, when routing networks are trained continually on multiple tasks we observe that the learned routing probabilities do not make use of all the experts. Typically we observe that the routing probabilities for task 1 converge to use some subset of the experts with high probability. When we start task 2, the router greedily chooses to use the same experts as task 1. The other experts have seen very few training examples and so are close to random initialization, this leads to the undesired solution that all tasks end up sharing the same set of experts.

To avoid this problem we need all the experts after task $T$ to be well initialized for quick learning on task $T+1$. If task $T+1$ is similar to task $T$ an overlapping set of experts can be used for both tasks and vice versa. We propose the \textit{co-training} method, see Algorithm \ref{algo:method}. For co-training, a data structure is maintained recording all experts that have been used in \textit{any} previous or current task. Supposing a maximum of $k$ experts can be activated in a given layer, an expert is considered \textit{used} on a task, if the learned routing probability for that expert on that task is in the top $k$ routing probabilities for that task. If we are currently training on a batch of examples from task $T$, then after taking the standard gradient descent step on the batch with the activated experts by the routers, we take an additional training step. In the additional training step we activate all \textit{unused} experts with equal weight, make a prediction for the same batch of examples from task $T$ and take a gradient descent step. Thus at task $T+1$, each expert has either been used on at least one previous task or has been trained via co-training on \textit{all} previous tasks, providing a diverse, well initialized set of experts for the router to choose from on the new task. Empirically we find that co-training helps reduce forgetting and improves average accuracy, see \S \ref{sec:experiments}.

\section{Experiments}
\label{sec:experiments}
We test our proposed method with routing networks and co-training on two standard continual learning benchmarks; MNIST-Rotations \cite{larochelle2007empirical} and MNIST-Permutations \cite{kirkpatrick2017overcoming}. 20 tasks are generated from the original MNIST dataset \cite{lecun2010mnist} via rotation and fixed random permutations of the pixels of the original image respectively. These tasks are then presented continually to the network.
We use two quantitative metrics popular in the continual learning literature to evaluate our method. \textit{Backward transfer} (\textbf{BWT}) measures to what extent future tasks impact the accuracy of past tasks and \textit{average accuracy} (\textbf{ACC}) is the mean accuracy over all tasks after completing training on all tasks, see \citet{lopez2017gradient} for definitions. 

\paragraph{Experimental Setup}
We compare against a state-of-the-art continual learning method, a shared bottom neural network with a small reservoir sampling episodic memory (maximum 1,000 examples stored) \cite{chaudhry2019continual}. The shared bottom network is a 2 hidden layer fully-connected neural network with 256 units in each hidden layer. All networks have task specific linear output heads with softmax activation function. The batch size is set to 10 and after each batch an additional batch of 10 examples is drawn from the episodic memory. Our method uses a mixture-of-experts routing network, with 2 hidden layers. Each hidden layer has 20 experts and $k = 4$ (maximum 4 experts active per layer). Each expert is a single fully-connected layer. For fair comparison we restrict the number units in each expert so that each method has an \textit{equal number of parameters}. The MoE network also has access to a similar episodic memory.

\paragraph{Results}
\begin{figure}[t]
% \vskip 0.2in
\begin{center}
\includegraphics[width=0.67\columnwidth]{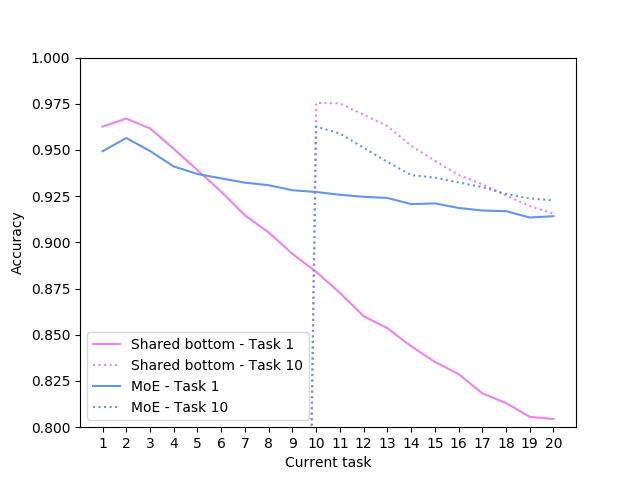}
\includegraphics[width=0.67\columnwidth]{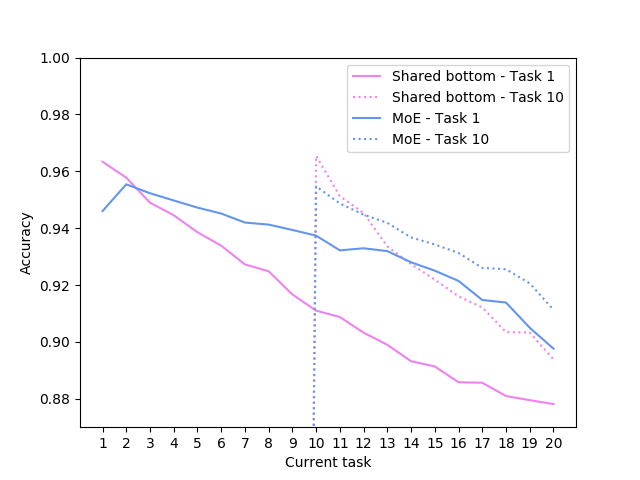}
\caption{Task 1 and 10 test accuracies evaluated after training on each task for MNIST-Rot (top) and MNIST-Perm (bottom). Both methods have access to a replay buffer of size 1,000 examples. 
MoE method is trained with co-training.}
\label{fig:accuracy_graph_mnist_rot}
\end{center}
\vskip -0.2in
\end{figure}

\begin{figure}[t]
% \vskip 0.2in
\begin{center}
\centerline{\includegraphics[width=0.7\columnwidth]{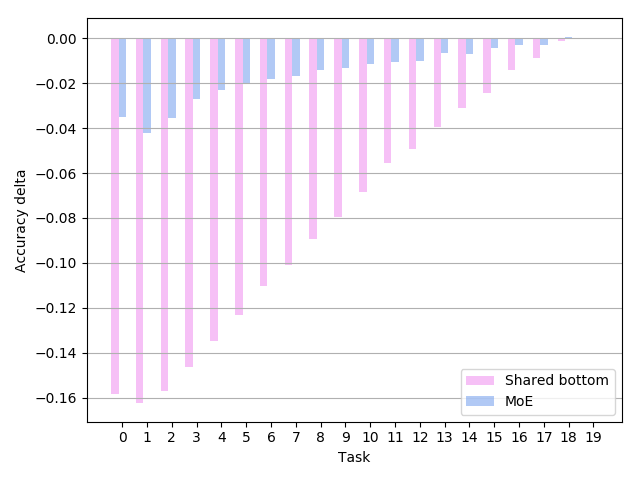}}
\centerline{\includegraphics[width=0.7\columnwidth]{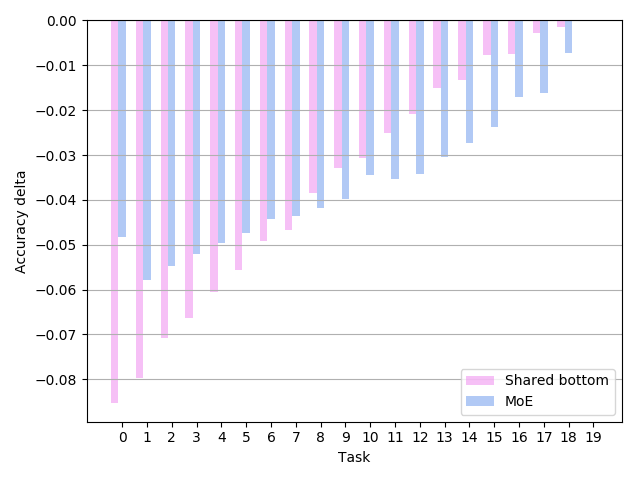}}
\caption{Difference in accuracy for each task, immediately after training on the task vs.\ at the end of training on all tasks, for MNIST-Rot (top) and MNIST-Perm (bottom). Both methods have access to a replay buffer of size 1,000 examples. MoE method is trained with co-training.}
\label{fig:accuracy_delta_mnist_rot}
\end{center}
\vskip -0.3in
\end{figure}

In Table \ref{tab:results} we see that the MoE network with co-training suffers less negative BWT and has higher average accuracy on both MNIST benchmarks when compared to the shared bottom architecture. We further see that the proposed co-training method improves both BWT and ACC. We also present plots of the task 1 and 10 accuracies over time, see Fig. \ref{fig:accuracy_graph_mnist_rot}. We see that for both benchmarks the MoE network is slightly slower to learn new tasks, perhaps due to the stochasticity in the learning process reducing sample efficiency. However the MoE network is much more robust to forgetting, with the accuracy curves significantly flatter than for the shared bottom architecture. Figure \ref{fig:accuracy_delta_mnist_rot} shows the reduction in accuracy for each task from immediately after learning the task to the end of training. We see that the routing networks suffer much less forgetting of early tasks than the shared bottom network.
This demonstrates empirically the claimed advantage of routing networks for continual learning, namely that when future tasks use different sets of experts to previously learned tasks, then the previously learned knowledge is protected from interference.

%\begin{figure}[ht]
%\vskip 0.2in
%\begin{center}
%\centerline{\includegraphics[width=\columnwidth]{test_accuracies_over_time_%MNIST_Perm.png}}
%\caption{Task 1 and 10 test accuracies evaluated after training on each task for MNIST-Perm. Both methods have access to a replay buffer of size 1,000 examples. MoE method is trained with co-training.}
%\label{fig:accuracy_graph_mnist_perm}
%\end{center}
%\vskip -0.2in
%\end{figure}

%\begin{figure}[ht]
%\vskip 0.2in
%\begin{center}
%\centerline{\includegraphics[width=\columnwidth]{accuracy_delta_per_task_MNIST_Perm.png}}
%\caption{Difference in accuracy for each task, immediately after training on the task vs.\ at the end of training on all tasks for MNIST-Perm. Both methods have access to a replay buffer of size 1,000 examples. MoE method is trained with co-training.}
%\label{fig:accuracy_delta_mnist_perm}
%\end{center}
%\vskip -0.2in
%\end{figure}

\paragraph{Interpretability of the learned routing}

\begin{figure}[t]
% \vskip 0.2in
\centering
\begin{subfigure}
    \centering
    \includegraphics[width=\columnwidth]{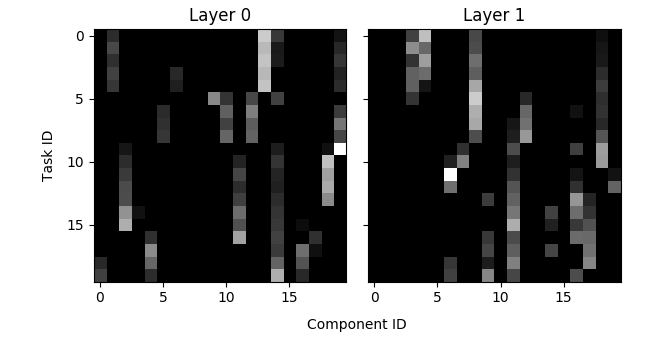}
    \vskip -0.2in
    \caption{Learned routing matrix for each layer on MNIST-Rot.}
    \label{fig:routing_matrix}
\end{subfigure}

\vskip 0.1in

\begin{subfigure}
    \centering
    \includegraphics[width=\columnwidth]{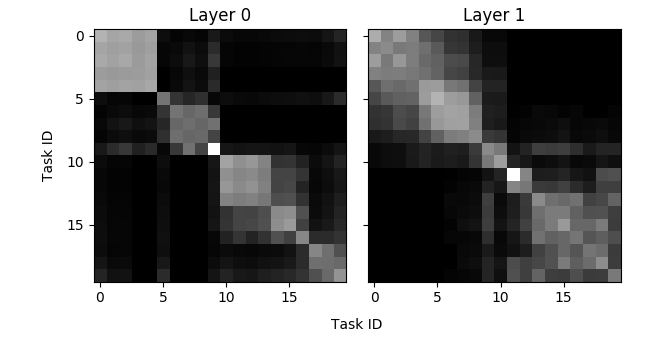}
    \vskip -0.2in
    \caption{Task similarity matrix for each layer on MNIST-Rot. Inferred from the routing matrix by matrix multiplication of the routing matrix with it's own transpose.}
    \label{fig:task_similarities}
\end{subfigure}
\label{fig:routing_probs}
\vskip -0.1in
\end{figure}

For MNIST-Rot, the tasks are presented in order of rotation, from 0 degrees rotation to 180 degrees rotation in increments of 9 degrees. Fig.\ \ref{fig:routing_matrix} shows the learned routing probability matrix for the two layers of the network for a single MNIST-Rot run. We see that the learned routing matrix is indeed sparse and that subsequent tasks often share some experts, but tasks far away from each other have mostly disjoint sets of experts, as one would expect.

Recall that the learned routing probability vector is conditioned on the task ID and hence can be understood as a task embedding. This allows us to embed tasks in a common space, compute distances between them and compare them.
Fig.\ \ref{fig:task_similarities} shows the inferred task similarity matrix, which is formed by the matrix multiplication of the routing matrix with its own transpose, such that the $(i, j)^{th}$ element contains the dot-product of the routing probability vector for the $i^{th}$ and $j^{th}$ tasks.
We see a block diagonal structure, indicating that tasks that are close in the degree of rotation share experts but tasks which have dissimilar levels of rotations tend not to share experts. This demonstrates that as anticipated the routing network has learned to group similar tasks together and thereby encourage positive transfer across similar tasks, while avoiding negative transfer/catastrophic forgetting from dissimilar tasks.

\section{Conclusion}

We have presented a new architectural solution for continual learning, sparse routing networks trained with co-training. The sparsity of the routing network ensures that gradients from different tasks only effect the weights of other tasks if the router has learned to group those tasks together. This promotes positive transfer across similar tasks and avoids negative transfer between dissimilar tasks. 
Our proposed architecture is general purpose and can be combined with existing and future methods in the continual learning literature. We have seen empirically that sparse routing networks trained with co-training and combined with a small episodic memory significantly reduces catastrophic forgetting and improves average accuracy across all tasks. However, we note that often the routing network's initial solution to a task has lower accuracy than the shared bottom method. We attribute this to lower sample efficiency in the routing networks, perhaps due to stochasticity in the routing decisions during training. This will be the focus of our future work.

\FloatBarrier

\clearpage

% Bring this back for the final version.
%\section*{Acknowledgements}
%We thank Basil Mustafa for providing Figure 1.

\bibliography{example_paper}
\bibliographystyle{icml2020}

\appendix
\section{Related Work}
\label{sec:related_work}

Conceptually, continual learning methods can be divided into the following categories: dynamic model
architectures, loss regularization, memory-based and methods that look into modeling the importance of weights.

\paragraph{Dynamic model architectures}
These methods add more capacity to the network as new tasks are ingested.
%added in order to be able to learn new tasks while not forgetting the knowledge of previous tasks.
Progressive networks \cite{rusu2016progressive} grow the network architecture as new tasks get ingested. When switching to a new task, the previous network parameters are kept frozen. Catastrophic forgetting is prevented by instantiating a new neural network for each task being solved, while transfer is enabled via lateral connections to features of previously learned networks.
On the other hand, Dynamically Expandable networks (DEN) \cite{DEN.2018} perform selective retraining and may choose to dynamically expand the network capacity upon arrival of a new task. If the new task cannot be sufficiently solved by the current network, this method expands the network by identifying drifting units, splitting/duplicating them and retraining them on the new task. Expert Gate \cite{aljundi2017expert} also adds a new network for each new task but uses an autoencoder to route examples to different experts at test time.

\paragraph{Loss regularization}
Some continual learning methods rely on modifying the loss function in order to cope with catastrophic forgetting.
For example, the MER method introduced in \cite{riemer2019learning} relies on two main ingredients: (a) a meta-experience replay buffer that stores samples from several tasks (using reservoir sampling) and (b) a loss function inspired by MAML (Model-Agnostic Meta-Learning) \cite{finn2017model} that takes into account the dot product between gradients at different points in order to maximize transfer and minimize interference.
Learning without Forgetting (LwF) \cite{LwF.2016} can be seen as a combination of distillation networks and fine-tuning.
This method essentially enforces that the output probabilities
for each sample in the new task be close to the produced output from the original network (when prompted with the same new data).
The PGMA method introduced by \citet{hu2019overcoming} splits the network parameters in two parts; one part that is shared among all tasks and another part that gets adapted to the test instance in order to classify it. The latter is generated by a neural network called the parameter generator. In order to cope with catastrophic forgetting, the PGMA method uses a notion similar to the distillation loss evaluated on replayed samples that are generated by another data generator network. %This loss term ensures that the output of the network does not change much when evaluated on the replayed samples; hence keeping the past learned knowledge.

\paragraph{Memory-based methods}
These methods store a small subset of the data seen so far in a  memory buffer and use this information later to avoid catastrophic forgetting. For example, Gradient Episodic Memory (GEM) \cite{lopez2017gradient} and A-GEM \cite{chaudhry2018efficient} uses this idea to ensure that the gradient on the current task does not interfere with gradients computed on samples stored in the memory. Interestingly, this method also allows for positive backward transfer i.e. improved performance on previously learned tasks.
A recent method called Orthogonal Gradient Descent \cite{OrthGradDescentCL.2019} proposes to enforce that the gradients for the new task are orthogonal to the gradients of the previous task. It modifies stochastic gradient descent such that the gradient direction is still a descent direction (i.e.\ loss gets lower) but is constrained to be orthogonal to the gradients of the previous task samples. The proposed method stores the gradients of previous tasks in a memory buffer and uses Gram-Schmidt orthogonalization to enforce the orthogonality.

\paragraph{Model weights importance}
These methods estimate the importance of individual weights to previous tasks and prevent them from being significantly modified when learning new tasks.
The core idea in \citet{UncertaintyGuidedCL.2020} is to represent the importance of the model weights by the inverse of their uncertainty, which allows modulation of the learning rate for each weight by its uncertainty. 
%This is done to ensure that the important weights will not be disturbed too much by learning new tasks. 
The uncertainty of each model weight is estimated using the Bayes-by-backprop (BBB) method \cite{blundell2015weight}.
Elastic Weight Consolidation (EWC) \cite{kirkpatrick2017overcoming} slows down learning on certain weights based on how important they are to previously seen tasks as measured by the Fisher information matrix.  When learning a new task, the weights are regularized to stay close to the weights trained on previous tasks, with the amount of the regularization on each weight proportional to the Fisher information for that weight.

\end{document}